\documentclass[12pt]{article}
\usepackage{sbc-template}
\usepackage{graphicx,url}
\usepackage[brazil]{babel}   
\usepackage[utf8]{inputenc}
\usepackage{graphicx}
\usepackage{subfigure}
\usepackage{amsmath}
\usepackage{bbm}
\usepackage{cite}
\usepackage{listings}
\usepackage{color}
\usepackage{indentfirst}
\usepackage{float}

\title{Uma implementação do jogo Pedra, Papel e Tesoura utilizando Visão Computacional}
\author{Ezequiel França dos Santos\inst{1}, Gabriel Fontenelle\inst{1}}
\address 
{Centro Universitário Senac - Campus Santo Amaro
  (SENAC-SP)\\
  Av. Engenheiro Eusébio Stevaux, 823 -- São Paulo -- CEP 04696-000 -- SP -- Brasil
  \email{{ezefranca.br,colecionador.gabriel},{(@gmail.com})}
}

\begin{document} 
\maketitle

\begin{abstract}
This paper presents a game, controlled by computer vision, in identification of hand gestures (hand-tracking). The proposed work is based on image segmentation and construction of a convex hull with Jarvis's Algorithm , and determination of the pattern based on the extraction of area characteristics in the convex hull.
\end{abstract}
     
\begin{resumo} 
Este trabalho apresenta um jogo, controlado através de visão computacional, na identificação de gestos da mão (hand-tracking). O trabalho proposto baseia-se na segmentação da imagem e construção de um fecho convexo com algoritmo de Jarvis e determinação do padrão com base na extração de características de sua área.
\end{resumo}

\section{Introdução}

A busca por meios que tornem os jogos mais interativos tem sido muito explorada. Muitos destes novos meios envolvem a área de visão computacional. Este trabalho apresenta um estudo sobre a viabilidade de utilização de uma webcam como dispositivo de interação baseado em gestos da mão, especificamente para o jogo Pedra, Papel e Tesoura. 

Neste trabalho, para o reconhecimento de gestos da mão utilizamos a combinação de algumas técnicas e para um reconhecimento satisfatório, fizemos um pré-processamento da imagem, passando a mesma para escala de cinza em seguida binarizando e por fim aplicamos um filtro para detecção de bordas. Em posse da imagem esqueletizada da mão, trabalhos o fecho convexo do conjunto de pontos e extraimos algumas características de sua área. 

\section{Desenvolvimento}

Primeiramente capturamos a imagem da camera, após isto fazemos a subtração de fundo com intuito de isolar melhor a mão para a captura de seus gestos. A próxima etapa foi a normalização da imagem para tons de cinza, seguida da sua binarização. Com a imagem binarizada, começamos o processo de reconhecimento, primeiramente reconhecedo as bordas da mão, e calculando o fecho convexo em torno da mesma. Em seguida, obtemos a determinação do padrão com base na extração de características da área do fecho convexo.

A figura 1, mostra a sequência adotada e cada etapa será brevemente descrita no decorrer deste trabalho.  

\begin{figure}[H]
\centering
\includegraphics[width=1\textwidth]{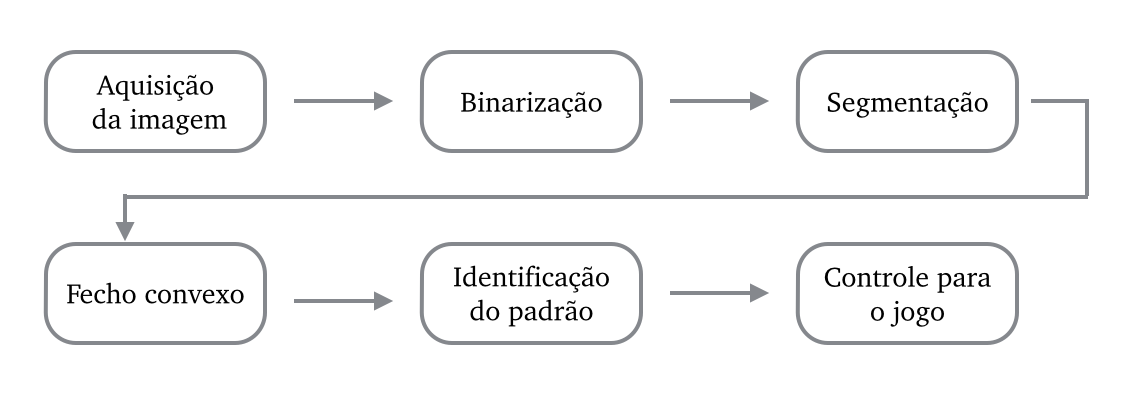}
\caption{Fluxo no processamento da imagem} \label{fig1}
\end{figure}

\subsection{Aquisição da imagem}
A biblioteca utilizada neste trabalho\cite{hashimoto}, permite o acesso a imagens de câmeras através da plataforma OpenCV. Ela permite trabalhar com a imagem como se a mesma fosse uma matriz tridimensional. A primeira dimensão da matriz representa a altura, a segunda dimensão representa a largura e a terceira representa os canais vermelho, verde, azul (RGB) e alfa ($\alpha$) de cada pixel.  O fator alfa de cada pixel é utilizado para determinar como as cores serão unidas quando imagens de cores diferentes estiverem sobrepostas.

\begin{figure}[H]
\centering
\includegraphics[width=.60\textwidth]{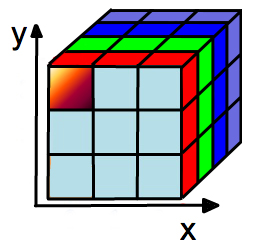}
\caption{Representação da matriz tridimensional da imagem de XY} \label{fig1}
\end{figure}

Os valores representam respectivamente a quantidade de vermelho (R), verde (G), azul (B) e alfa($\alpha$) do pixel na posição vertical (y) e posição horizontal (x). Cada quantidade, está entre 0 e 255.

\subsection{Normalização para escala de cinza}

A primeira fase de pré-processamento trata-se da normalização da imagem colorida para tons de cinza. A normalização foi feita com base no valor médio dos canais de cores da imagem, conforme a equação \ref{eq:normalizacao}.

\begin{equation} \label{eq:normalizacao}
Valor{_{cinza}}{_{i, j}}= \sum_{1}^{n} \left |\frac{R + G + B}{3}\right |{_{i, j}}
\end{equation}

Onde:

\begin{itemize}  
  \item $Valor{_{cinza}}$ -\; valor\,entre\,0\,-\,255\,para\,a\,escala\,de\,cinza
  \item $R $ -\; valor\,vermelho\,do\,ponto
  \item $G $-\; valor\,verde\,do\,ponto
  \item $B $-\; valor\,azul\,do\,ponto
  \item $n $-\; quantidade\,de\,pontos\,da\,imagem
  \item $i, j$ \,-\;coordenadas\, (x,y)\, do\, ponto\, na\, imagem
\end{itemize}

\begin{figure}[H]
\centering
\mbox{\subfigure{\includegraphics[width=3.2in]{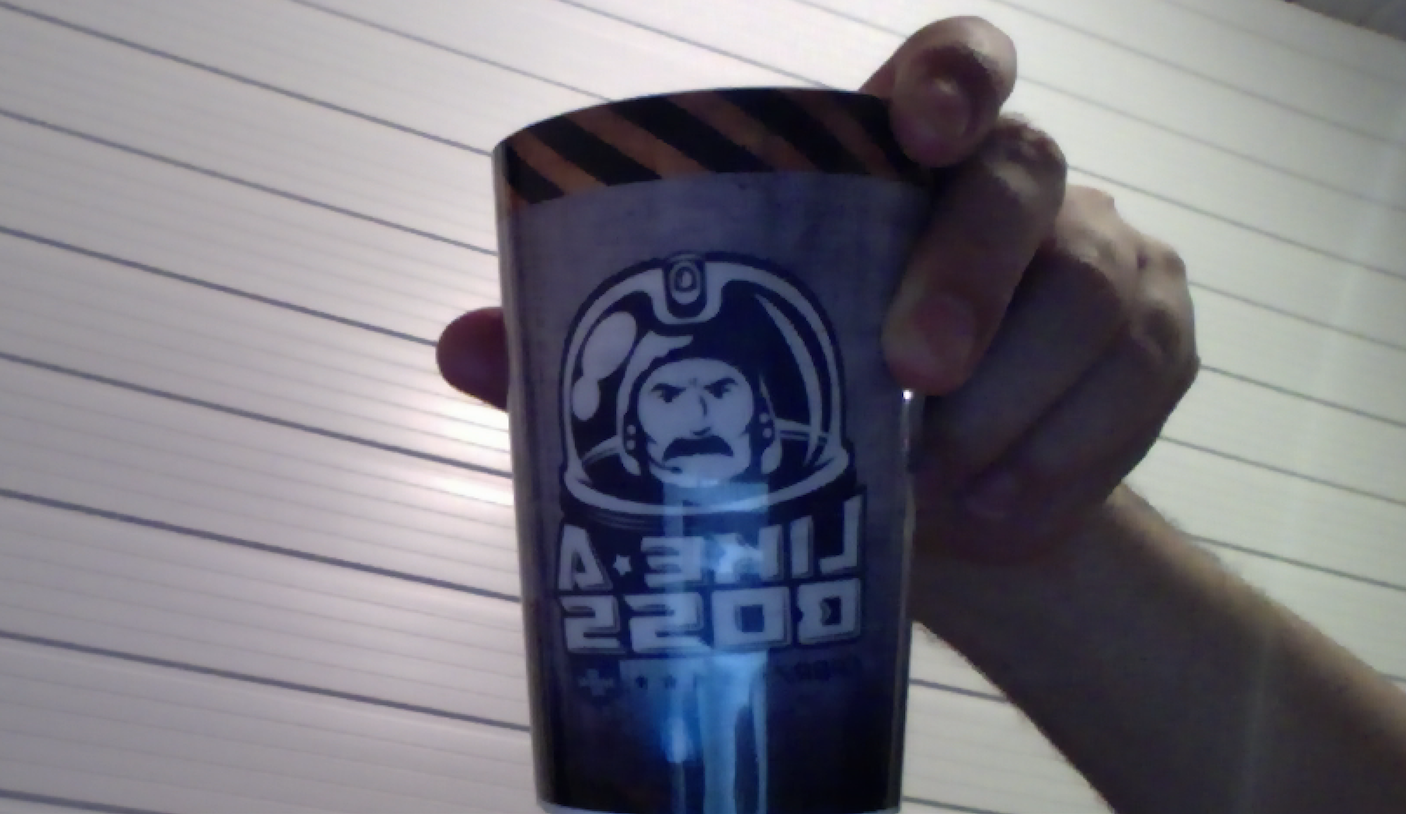}}\quad
\subfigure{\includegraphics[width=3.2in]{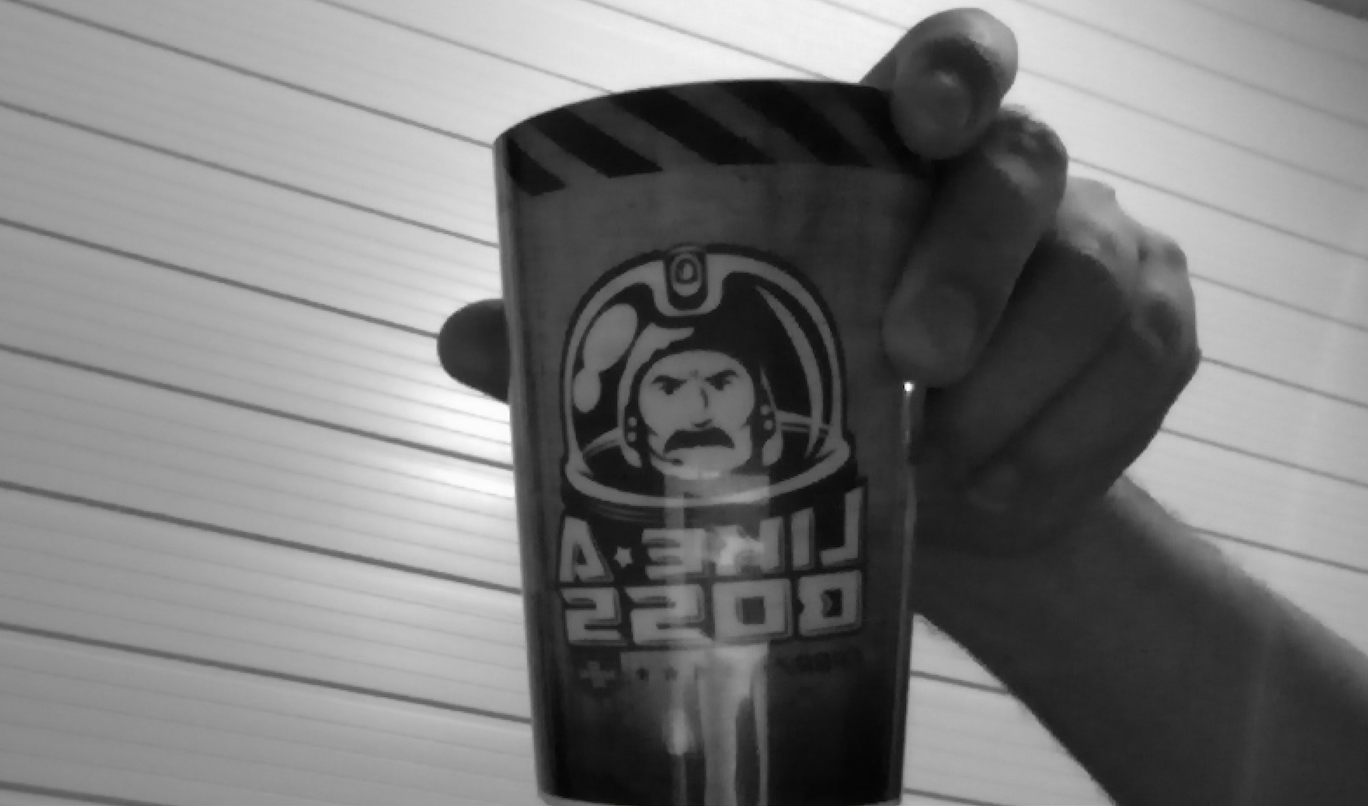} }}
\caption{Imagem normal (a) e imagem normalizada em cinza (b)} \label{fig2}
\end{figure}

A Figura \ref{fig2} apresenta o resultado deste processo.

\subsection{Binarização da imagem}

Existem diversos algoritmos para binarização de imagens, dentre a lista de soluções para este  o algoritmo de Otsu, por ser de fácil implementação e apresentar resultados satisfatórios nos experimentos realizados.

O método de Otsu é um método de \textit{thresholding} global, isto é, o valor obtido é
uma constante, para escolha do melhor limiar. A base deste método é sua interpretação
do histograma como uma função de densidade de probabilidade
discreta \cite{Limiar}, do seguinte modo:

\begin{equation}\label{eq:histograma_norm}
  p_r(r_q) = \frac{n_q}{n}, q = 0, 1, 2, ..., L-1
\end{equation}

Onde:

\begin{itemize}
  \item $ n $ é o total de \textit{pixels} da imagem;
  \item $ n_q $ é o total de \textit{piixels} que tem intensidade $ r_q $ e
  \item $ L $ é o total de níveis de intensidade na imagem.
\end{itemize}

O método de Otsu escolhe o limiar de valor $ k $, tal que $ k $ é um nível de
intensidade que divide o histograma em duas classes
$ C_0 = [0, 1, ..., k-1] $ e $ C_1 = [k, k+1, ..., L-1] $, e que maximise a
variância $ \sigma_{B}^2 $ definida como:

\begin{equation}\label{eq:maximizacao_variancia}
  \sigma_{B}^2 = \omega_0(\mu_0 - \mu_T)^2 + \omega_1(\mu_1 - \mu_T)^2
\end{equation}

Sendo:

\begin{subequations}\label{eq:somatorios_maximizacao}
\begin{align}
  \omega_0 = \sum_{q=0}^{k-1} p_q(r_q)\\
  \omega_1 = \sum_{q=k}^{L-1} p_q(r_q)\\
     \mu_0 = \sum_{q=0}^{k-1} \frac{qp_q(r_q)}{\omega_0}\\
     \mu_1 = \sum_{q=k}^{L-1} \frac{qp_q(r_q)}{\omega_1}\\
     \mu_T = \sum_{q=0}^{L-1} qp_q(r_q)
\end{align}
\end{subequations}

\begin{figure}[H]
\centering
\mbox{\subfigure{\includegraphics[width=3.2in]{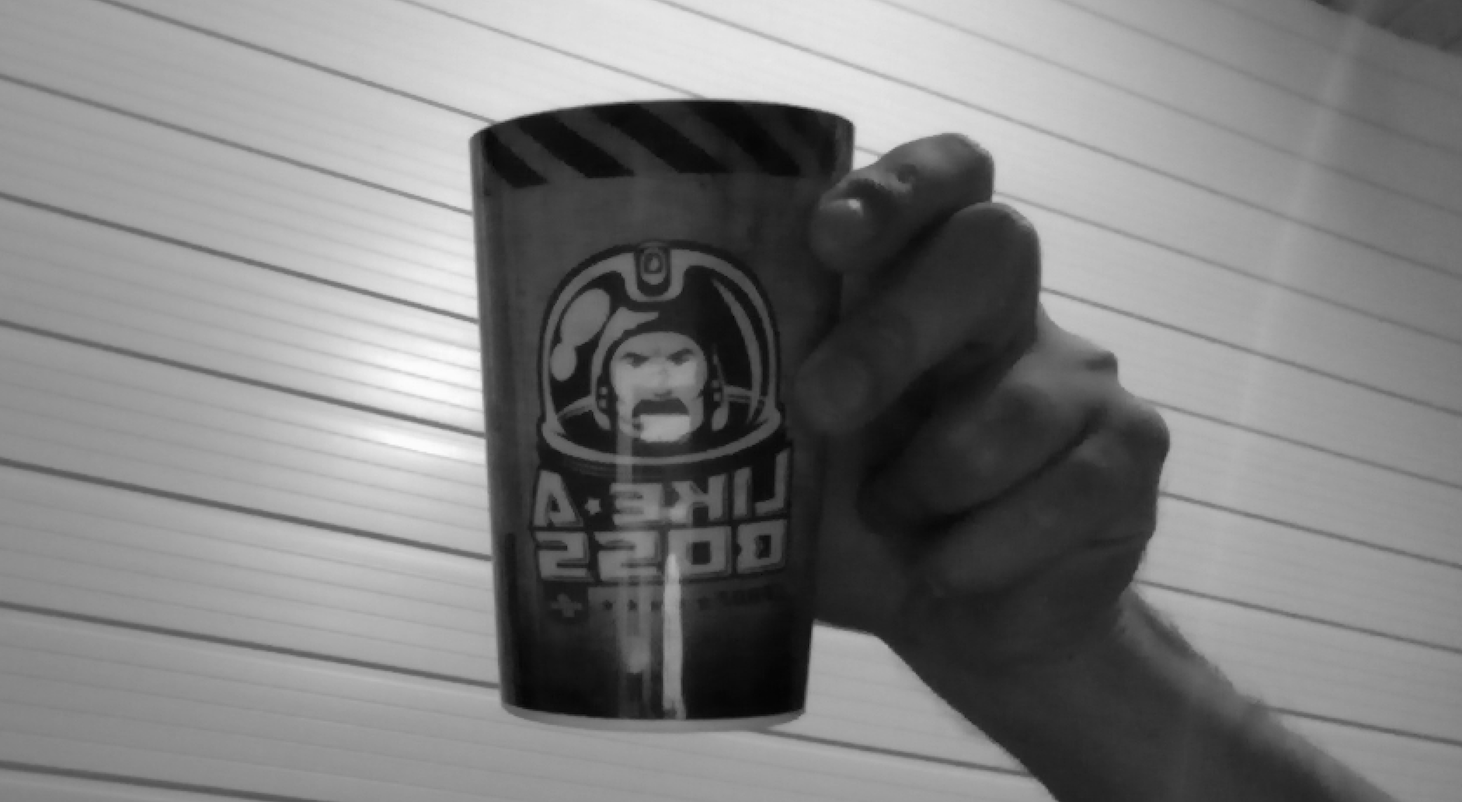}}\quad
\subfigure{\includegraphics[width=3.2in]{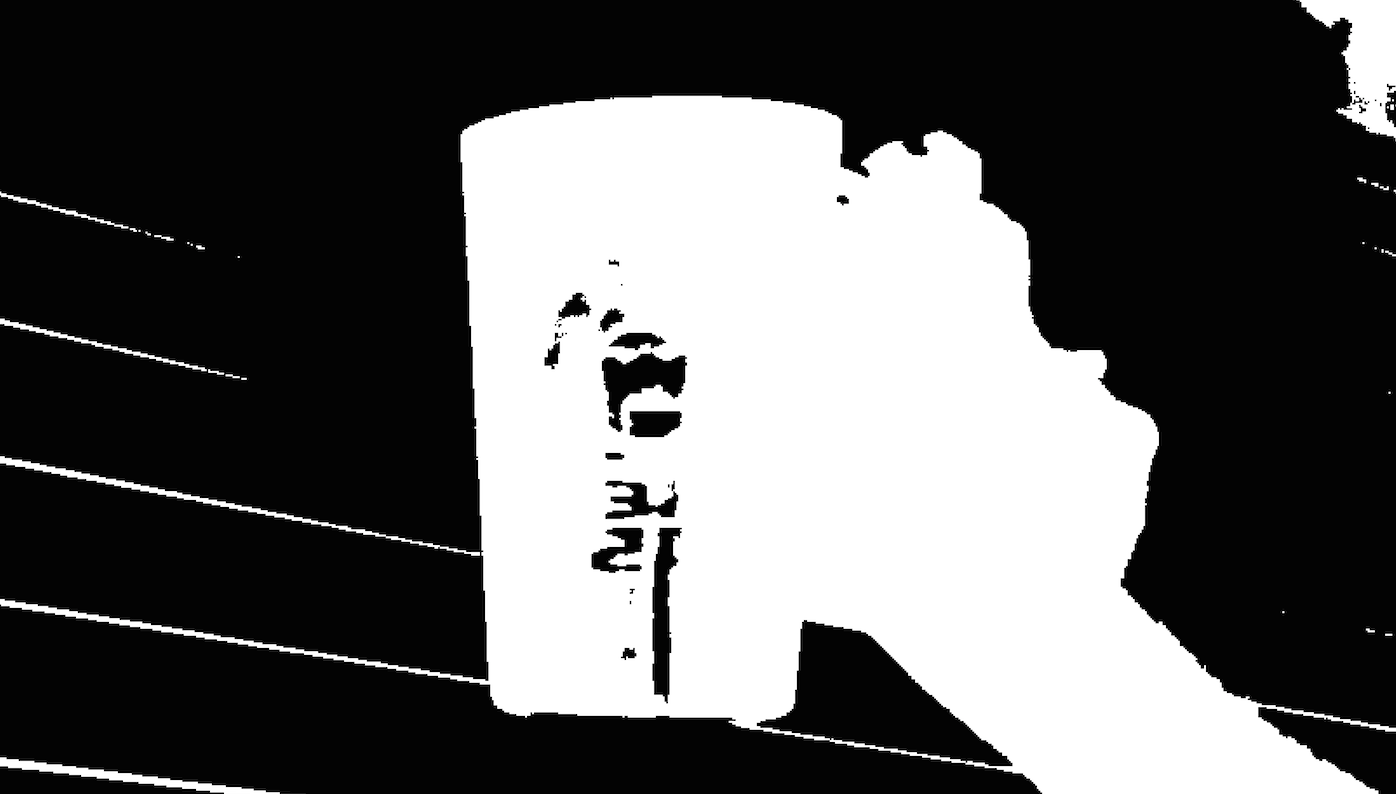} }}
\caption{ Imagem binarizada com limiar definito pelo método de Otsu (b). }
\label{fig:bin_otsu}
\end{figure}

O resultado da binarização com limiar ajustado segundo o método de Otsu pode
ser observado na Figura \ref{fig:bin_otsu}

\subsection{Remoção de fundo}

A subtração de fundo foi implementada utilizando uma técnica de subtração
simples. Os valores do primeiro frame são comparados e se a diferença for maior que um determinado threshold com o frame atual, o pixel é considerado como objeto, caso contrário, será considerado fundo. Os resultados dessa técnica não são extremamente eficazes, pois não levam em consideração nenhum embasamento estatístico e nem uma etapa de treinamento, mas para este trabalho mostraram-se suficientes ajustando-se o $Threshold$ (\ref{eq:threshold}) de acordo com o ambiente.

\begin{subequations} \label{eq:r}
\begin{align}
r{_{i, j}} = \left|R{_{primeiro}} - R{_{atual}}\right |{_{i,j}} 
\\
g{_{i, j}} = \left|G{_{primeiro}} - G{_{atual}}\right |{_{i,j}} 
\\
b{_{i, j}} = \left|B{_{primeiro}} - B{_{atual}}\right |{_{i,j}} 
\end{align}
\end{subequations}

\begin{equation} \label{eq:threshold}
Threshold = r{_{i, j}} + g{_{i, j}} + b{_{i, j}}
\end{equation}

Onde:

\begin{itemize}  
  \item $Threshold -\; valor\,entre\,0\,-\,255\,para\,a\,escala\,de\,cinza $
  \item $R{_{primeiro}} -\; valor\,vermelho\,do\,ponto\,no\,primeiro\,frame\,$
  \item $G{_{primeiro}} -\; valor\,verde\,do\,ponto\,no\,primeiro\,frame \,$
  \item $B{_{primeiro}} -\; valor\,azul\,do\,do\,ponto\,no\,primeiro\,frame\,$
  \item $R{_{atual}}    -\; valor\,vermelho\,do\,ponto\,no\,frame\,atual$
  \item $G{_{atual}}    -\; valor\,verde\,do\,ponto\,no\,frame\,atual$
  \item $B{_{atual}}    -\; valor\,azul\,do\,do\,ponto\,no\,frame\,atual$
  \item $(i, j)       \,-\;coordenadas\, (x,y)\, do\, ponto\, na\, imagem$
\end{itemize}

E por fim:
\begin{equation}
(R, G, B){_{atual}} = 255 \rightarrow Threshold \geq K \wedge (R, G, B){_{atual}} = 0 \rightarrow Threshold < K
\end{equation}
Onde:

\begin{itemize}  
  \item $K -\; valor\,ajustado\,manualmente\,no\,programa\,de\,acordo\,com\,o\,ambiente$
\end{itemize}

\subsubsection{Detecção de bordas com filtro Sobel}

O filtro Sobel calcula o gradiente da intensidade da imagem em cada ponto, dando a direcção da maior variação de claro para escuro e a quantidade de variação nessa direcção, através de duas matrizes 3x3, que são convoluídas com a imagem original para calcular aproximações das derivadas - uma para as variações horizontais $Gx$ e uma para as verticais $Gy$.
\begin{center}{Máscara de Sobel 3x3}
$$
Gx=\left[\begin{array}{rrr}
-1&0&+1\\
-2&0&+2 \\
-1&0&+1
\end{array}\right]\quad
Gy=\left[\begin{array}{ccc}
-1&-2&-1\\
0& 0& 0 \\
+1&+2&+1
\end{array}\right]
$$
\end{center}
A magnitude do gradiente é dado por:

$$
|G|=\sqrt{Gx^2 + Gy^2}
$$	

\begin{figure}[H]
\centering
\mbox{\subfigure{\includegraphics[width=3.2in]{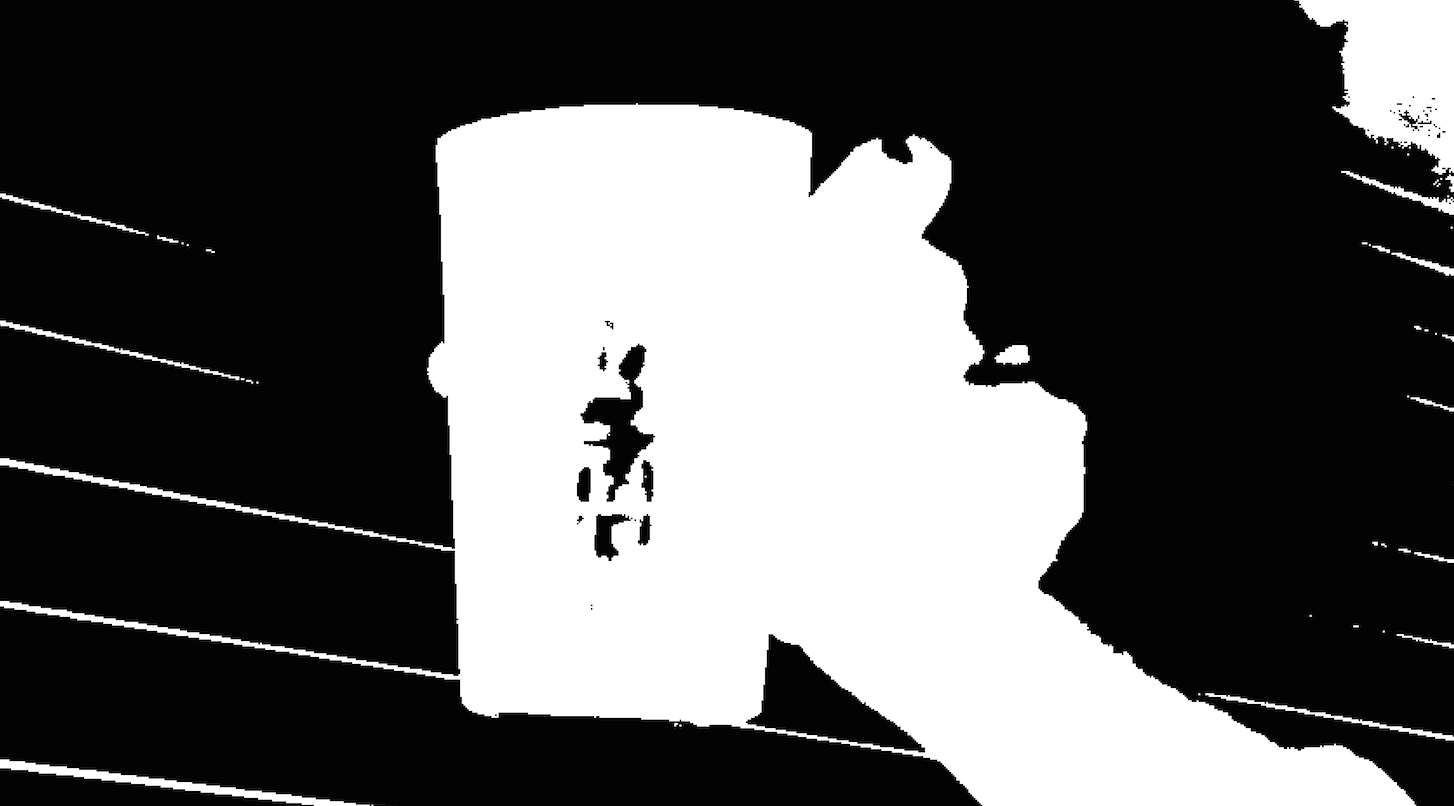}}\quad
\subfigure{\includegraphics[width=3.2in]{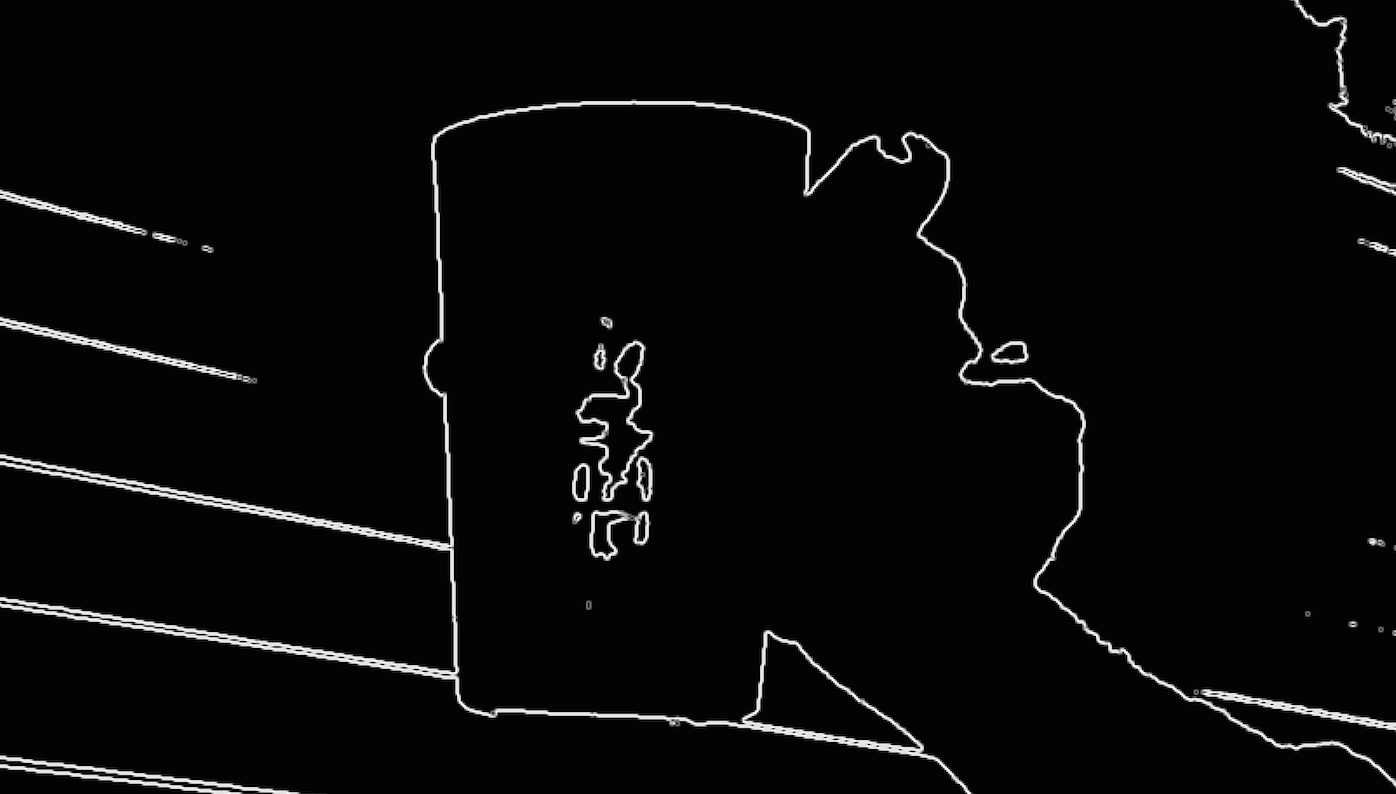} }}
\caption{ Resultado do filtro Sobel (b) e uma imagem binarizada (a).}
\label{fig:remove_fundo}
\end{figure}

\subsection{Reconhecimento dos padrões}
\subsubsection{Determinação do fecho convexo}
\begin{itemize}

\item Um conjunto ${\cal S} \subset {\cal R}^d$ é {\em convexo} se $\lambda
z_1 + (1-\lambda)z_2 \in {\cal S}$ sempre $z_1, z_2 \in{\cal S}$,
e $0 \le \lambda \le 1$. Resumidamente, ${\cal S}$ contém todos os segmentos de linha que conectam pares de pontos em ${\cal S}$.

\item O {\em fecho convexo} gerado por um conjunto de pontos ${\cal P}$ é
a intersecção de todos conjuntos convexos ${\cal S}$ que contem ${\cal P}$.
Se ${\cal P} = \{z_i\in{\cal R}^d, i=1\ldots, n\}$ é finito, pode ser expresso da seguinte forma:

$${\cal S} = \{ \sum_{i=1}^n \lambda_i z_i \;|\; 0 \le \lambda_i \le
1, \; \sum \lambda_i = 1\}.$$

\begin{center}
\scalebox{.6}{\includegraphics{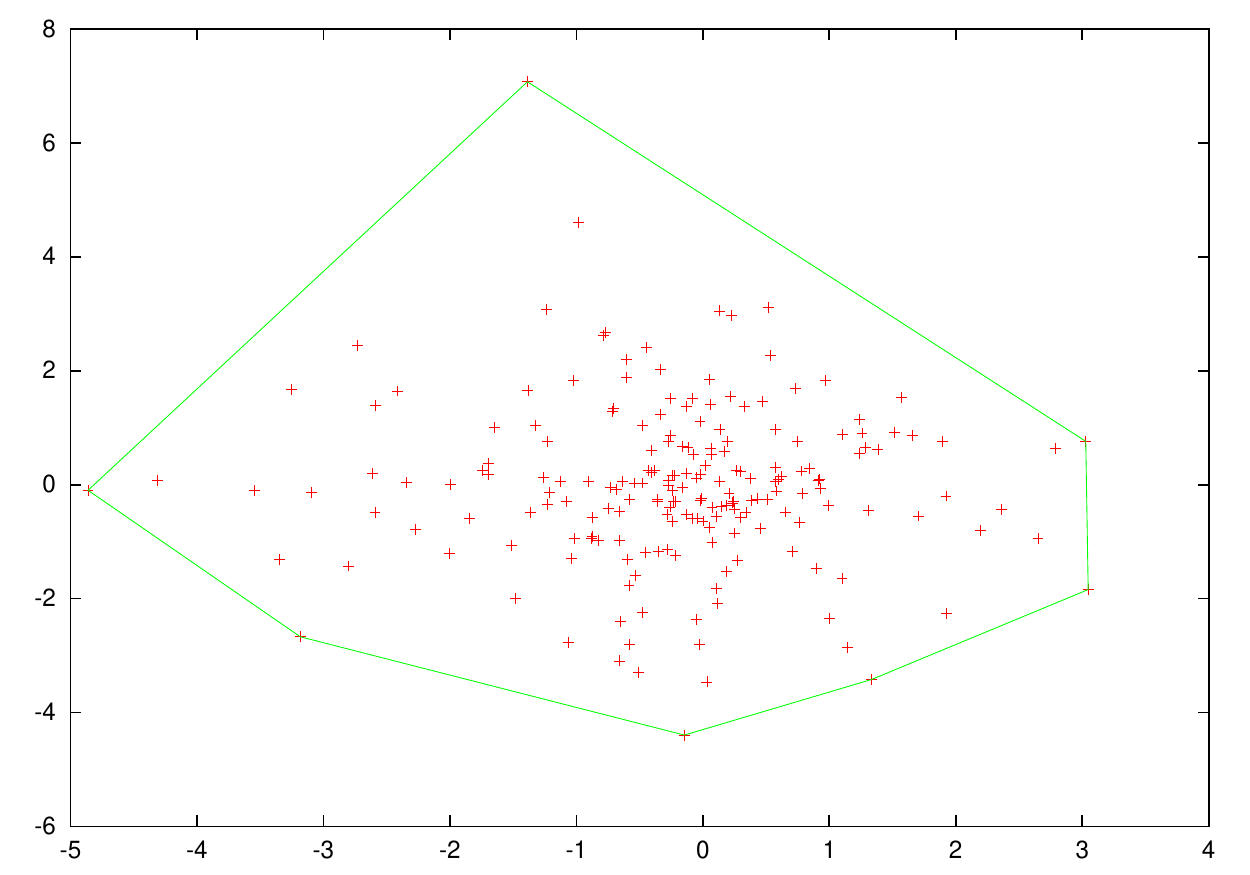}}
\end{center}

\item Se ${\cal P}$ é finito, existe um único subconjunto ${\cal
P}^* \subset {\cal P}$ de tamanho mínimo de tal modo que o fecho convexo de
${\cal P}^*$ é indentico ao fecho convexo de ${\cal P}$.  O conjunto
${\cal P}^*$ é chamado de conjunto {\em gerador de fecho convexo}.

\item Encontrar o conjunto de gerador do fecho convexo de um conjunto finito ponto
${\cal P}$ é um problema computacional difícil quando a dimensão $d$
é maior que $2$.  Se a $d = 2$ existem vários algoritmos eficientes.
\end{itemize}

\bigskip
\subsubsection{Algortimo da embrulho de presente}

\begin{itemize}

\item O {\em algoritmo da marcha de Jarvis}, popularmente conhecido como {\em gift
wrapping algorithm / algoritmo do embrulho de presente} , visita os pontos do fecho convexo de maneira ordenada.

\begin{enumerate}

\item Começamos com qualquer ponto do fecho. O ponto com maior coordenada em x é uma escolha natural. Chamamos esse ponto de $(X_0, Y_0)$.

\item Varremos ("marchamos") através de todos os pontos $(X_i, Y_i)$ e localizamos o ponto tal que o ângulo a partir da coordenada $(1,0)$ para $(X_i - X_0, Y_i - Y_0)$ é minímo.
Este é o próximo ponto de sentido anti-horário a partir de $(X_0,Y_0)$ no fecho, chamamos-o de $(X_1, Y_1)$.

\item Suponhamos que tenhamos localizado o ponto $(X_i,Y_i)$, $i=1, \ldots, m$ que ocorrem no sentido anti-horário ao fecho onde $m \ge 2$.
Calculamos todos os ângulos entre os vetores $(X_i - X_m, Y_i - Y_m)$ e
$(X_{m-1} - X_m, Y_{m-1} - Y_m)$, e procuramos o ponto $i$ que tenha o menor âgulo positivo.  Adicionamos este ponto ao fecho.

\item Retornarmos a etapa 3 até que $(X_m,Y_m) = (X_0,Y_0)$.

\end{enumerate}

\begin{center}
\scalebox{.8}{\includegraphics{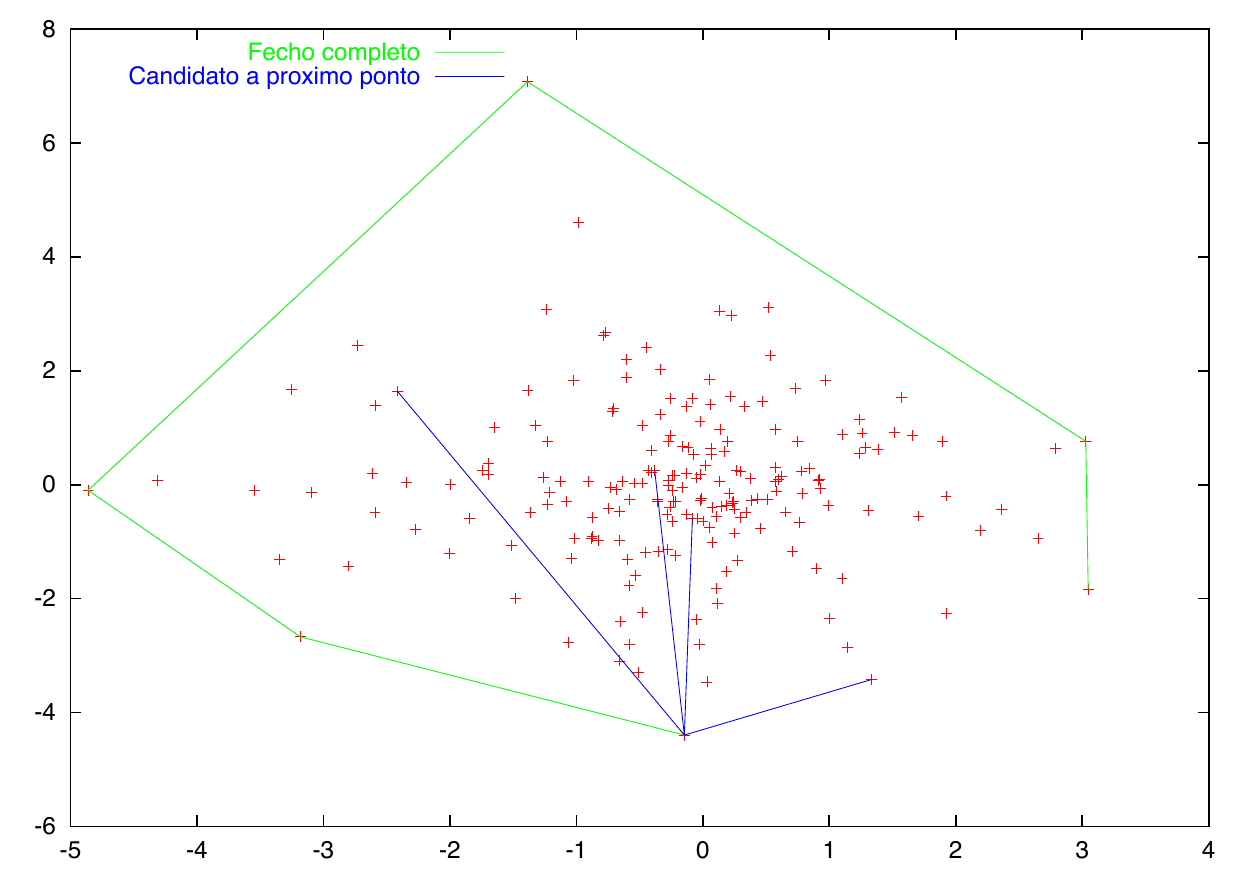}}
\end{center}
\end{itemize}

O algoritmo de marcha Jarvis tem no pior caso complexidade O($n^2$), o que ocorre se todos os pontos estão no fecho. Em geral, se $h$ pontos estão no fecho, a complexidade é O($nh$).

\begin{figure}[H]
\centering
\mbox{\subfigure{\includegraphics[width=3.2in]{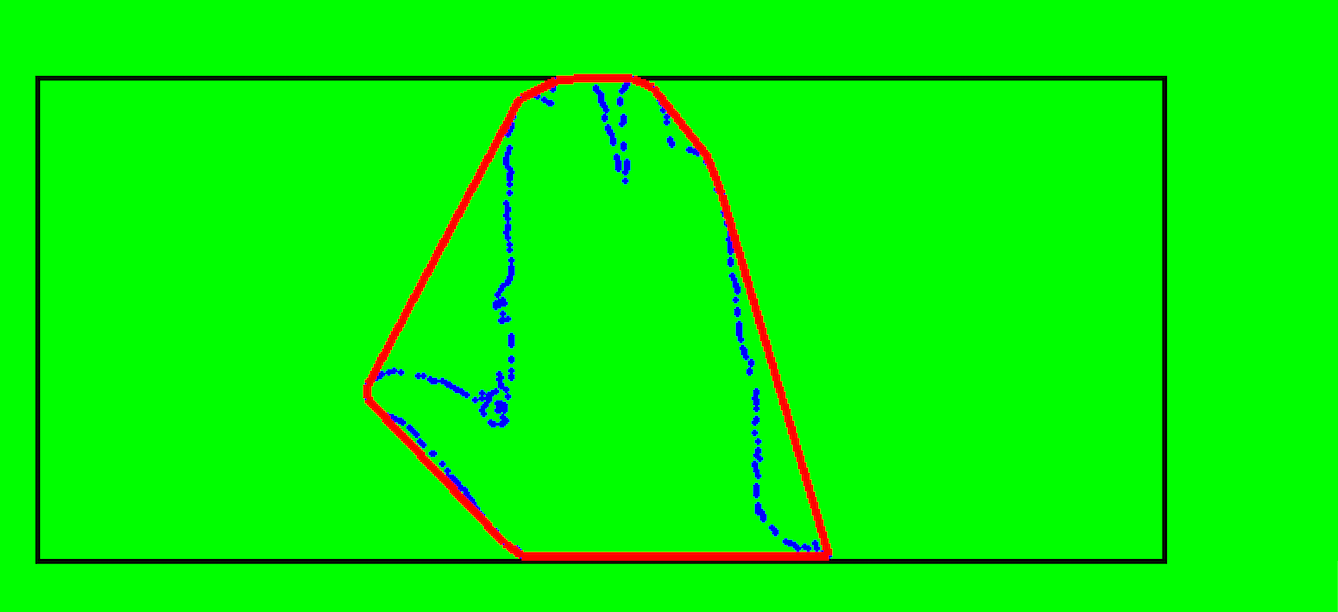}}\quad
\subfigure{\includegraphics[width=3.2in]{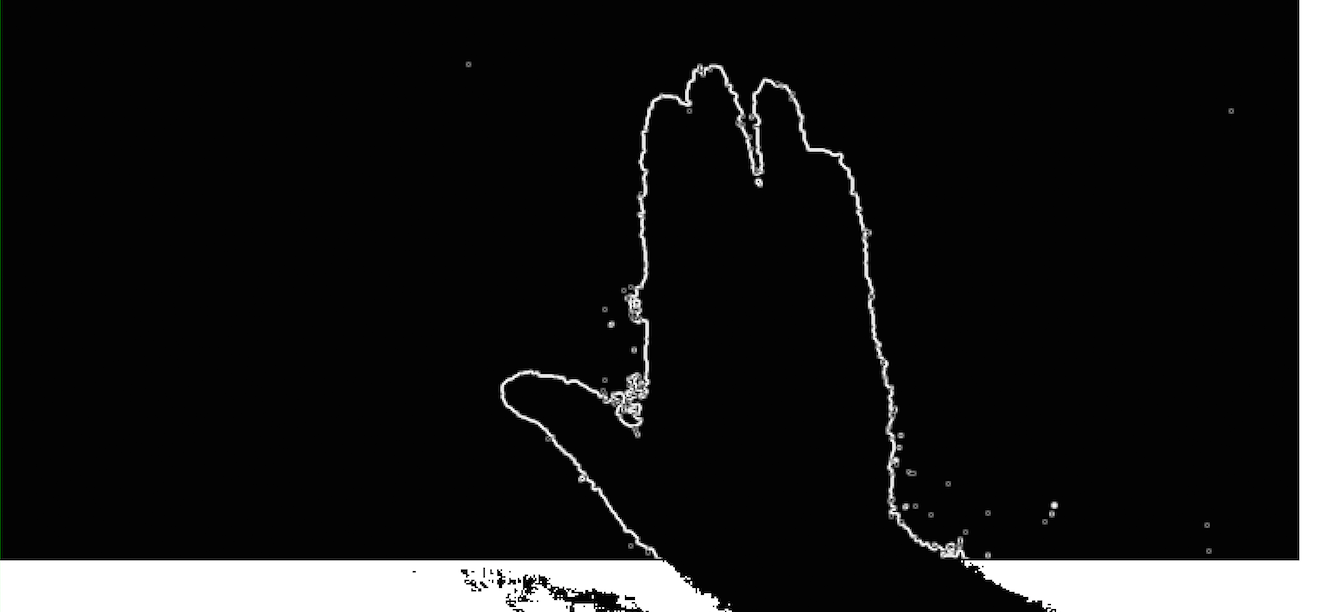} }}
\caption{Determinação do fecho convexo (a) na posição "papel"}
\label{fecho2}
\end{figure}

\subsection{Determinação do padrão}
\subsubsection{Cálculo da área do fecho convexo}

Utilizamos o método $Shoelace$ de Gauss para calcular a área do total do fecho convexo, obtendo com precisão seus pontos extremos. Com o método é possível obter a área fechada de qualquer região poligonal conhecendo apenas as coordenadas de seus vértices. A formula geral pode ser expressa como:

\begin{figure}[H]
\centering
\includegraphics[width=0.7\textwidth]{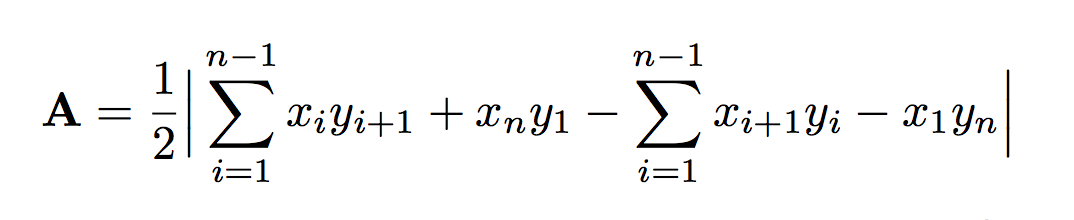}
\end{figure}

Onde:

\begin{itemize}  
  \item $A -\; area$
  \item $n -\; quantidade\, de\, vertices\, do\, poligono\,$
  \item $(x, y)\,-\;coordenadas\, (x,y)\, do\, ponto\,$
\end{itemize}

O método funciona em qualquer região poligonal, independentemente do número de lados. A razão pela qual esta fórmula é chamada a fórmula do cardaço de sapato é devido ao método utilizado.

Para determinarmos a área de um triângulo com vértices $(2,4)$, $(3,-8)$ e $(1,2)$, devemos construir a matriz por "andando" nos vértices do triângulo, terminando com o ponto que começamos.

\begin{figure}[H]
\centering
\includegraphics[width=0.25\textwidth]{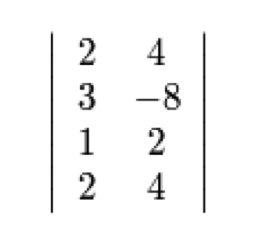}
\end{figure}

\begin{figure}[H]
\centering
\includegraphics[width=0.30\textwidth]{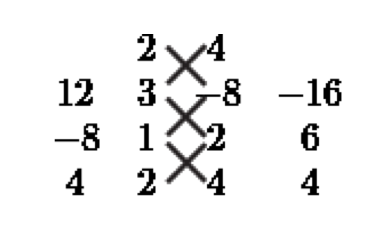}
\end{figure}

Funcionamente temos, multiplicamos através das linhas, em seguida, adicionamos os dois lados. Ficamos 8 e -6. Finalmente aplicamos na na fórmula:

\begin{figure}[H]
\centering
\includegraphics[width=0.5\textwidth]{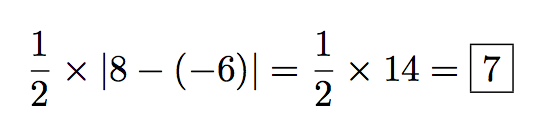}
\end{figure}

\subsubsection{Cálculo da área "branca"}

\begin{figure}[H]
\centering
\mbox{\subfigure{\includegraphics[width=3.2in]{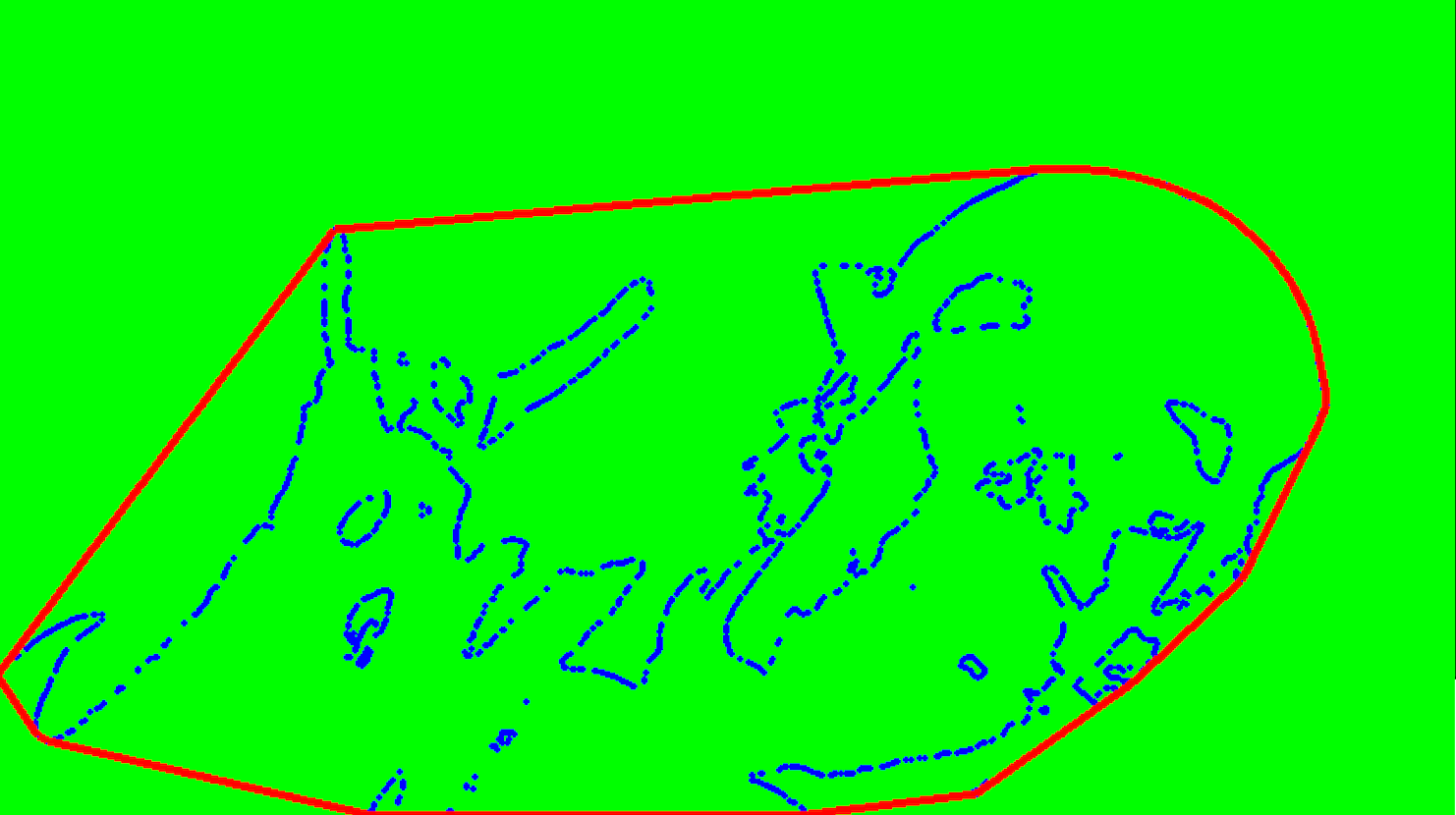}}\quad
\subfigure{\includegraphics[width=3.2in]{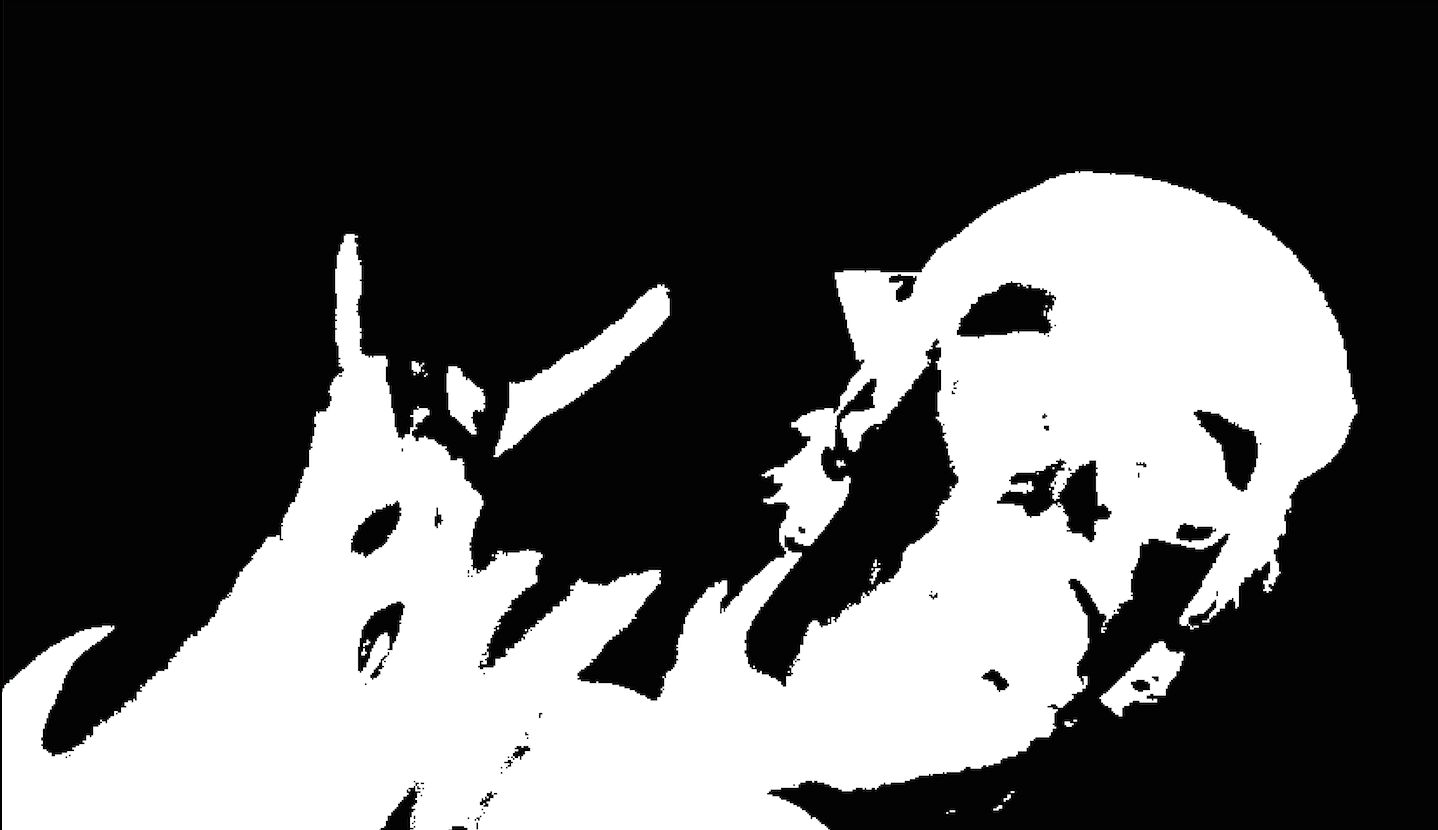} }}
\caption{ Fecho convexo (a) e imagem binarizada (b) utilizada no cálculo.}
\label{fig6}
\end{figure}

Para o cálculo da área "branca" foi utilizado uma cópia da matriz binarizada, sem aplicação do filtro de Sobel, e cada pixel dos vertices do fecho é gravado nessa cópia, e em seguida são verificados se os pixels dentro do fecho são "pretos" ou "brancos", sendo os pixels "brancos" contabilizados. A Figura \ref{fig6} demonstra as áreas em questão.

\subsection{Funcionalidades do jogo}

\subsubsection{Calibragem inicial}

Extraindo características do fecho convexo, como área total, área dos pixels "brancos" e pontos mínimo e máximo conseguimos determinar e diferenciar o gesto da mão do jogador.

Através da razão da área de pixels "brancos" para área total no fecho conseguimos diferenciar o gesto que representa a tesoura dos gestos "papel" e "pedra". Porém a diferenciação entre "papel" e "pedra" utilizando apenas a razão se mostrou inadequada, com ambos os gestos possuindo uma razão próxima a 1 no fecho. Para evitar situações em que "pedra" e "papel" são reconhecidos incorretamente é necessário uma calibragem inicial na qual é detectada e armazenada a distância do pontos mínimo e máximo do gesto "pedra", essa distância é então comparada em situações em que o gesto não é tesoura para determinar se é "pedra" ou "papel".

\begin{figure}[H]
\centering
\includegraphics[width=.60\textwidth]{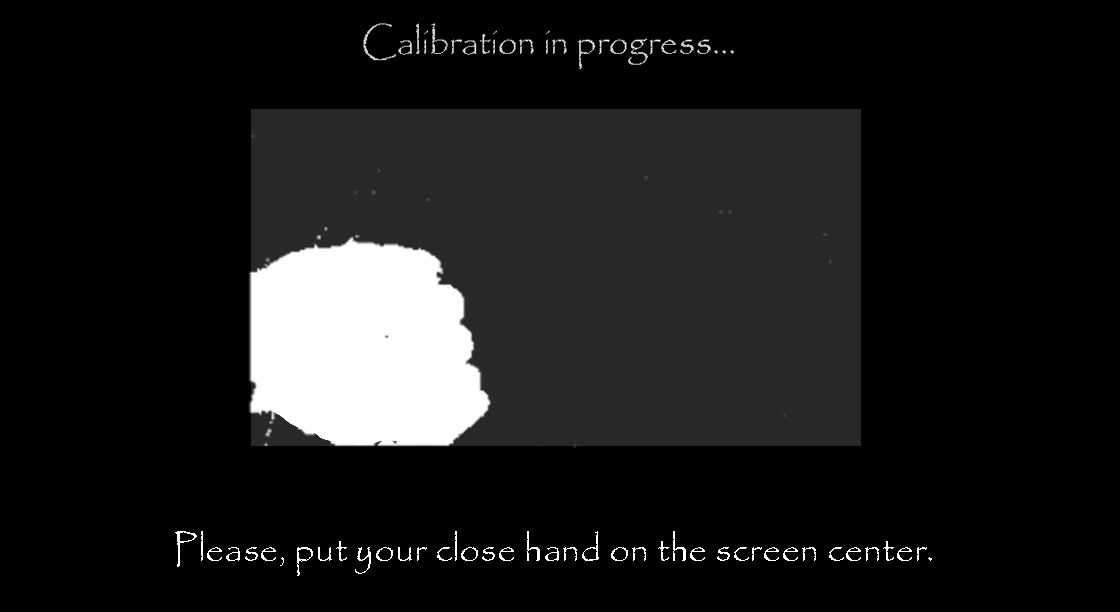}
\caption{Tela de calibragem inicial do jogo.} \label{fig1}
\end{figure}

\subsubsection{Multilinguagem}

Para possibilitar o uso de múltiplos idiomas no jogo foi utilizado arquivo personalizado para armazenar os textos de cada idioma e $Hash\ Table$. Cada arquivo de texto contém todas as frases que aparecem no jogo em um determinado idioma, o nome do arquivo recebe a sigla do idioma e país, como exemplo temos o idioma Português do Brasil a qual o arquivo com os textos recebe o nome de pt\_BR.conf, a extensão do arquivo para leitura do mesmo no jogo não é significante e foi escolhida apenas por ser utilizada em arquivos de configuração de servidores.

O conteúdo do arquivo possui a seguinte estrutura interna:\\

\centerline{Chave única que represente o texto=Texto a ser exibido}
\centerline{Segunda chave única que represente o texto=Segundo Texto}

Para facilitar $debug$ e organização do código a chave única é sempre o texto original no idioma inglês, nosso algoritmo na ausência do texto traduzido retorna a chave única. Com essa estrutura cada frase apenas pode ocupar uma única linha do arquivo de idioma.

Durante o carregamento do jogo o arquivo de texto do idioma previamente configurado como padrão é aberto e seu conteúdo é armazenado em uma $Hash\ Table$ contendo 24 índices. O número de índices foi escolhido por representar a quantidade de caracteres do alfabeto latino.

Para inserção e retorno de frases na Tabela Hash é gerado uma chave de índice utilizando a primeira letra da frase.

\begin{equation} 
\acute{\imath}ndice = c\ \%\  24
\end{equation}

Índice = representação numérica do primeiro caractere da frase \% (módulo) 24

Onde:

\begin{itemize}  
  \item $\acute{\imath}ndice -\; valor\,entre\,0\,-\,23 $
  \item $c -\; valor\,n\acute{u}merico\,que\,representa\,o\,primeiro\,caractere\,da\,chave$
  \item $\% -\; m\acute{o}dulo$
\end{itemize}

Nosso jogo apresenta uma quantidade pequena de frases, mas ainda há a probabilidade de ocorrer colisões com frases obtendo um mesmo índice. A fim de tratar essas colisões utilizamos um algoritmo simples, usamos um vetor de ponteiros que armazena as structs representando a chave única e o texto a ser exibido, quando ocorre colisões num determinado índice é utilizado lista ligada para armazenas as colisões do índice.

\begin{figure}[H]
\centering
\includegraphics[width=.80\textwidth]{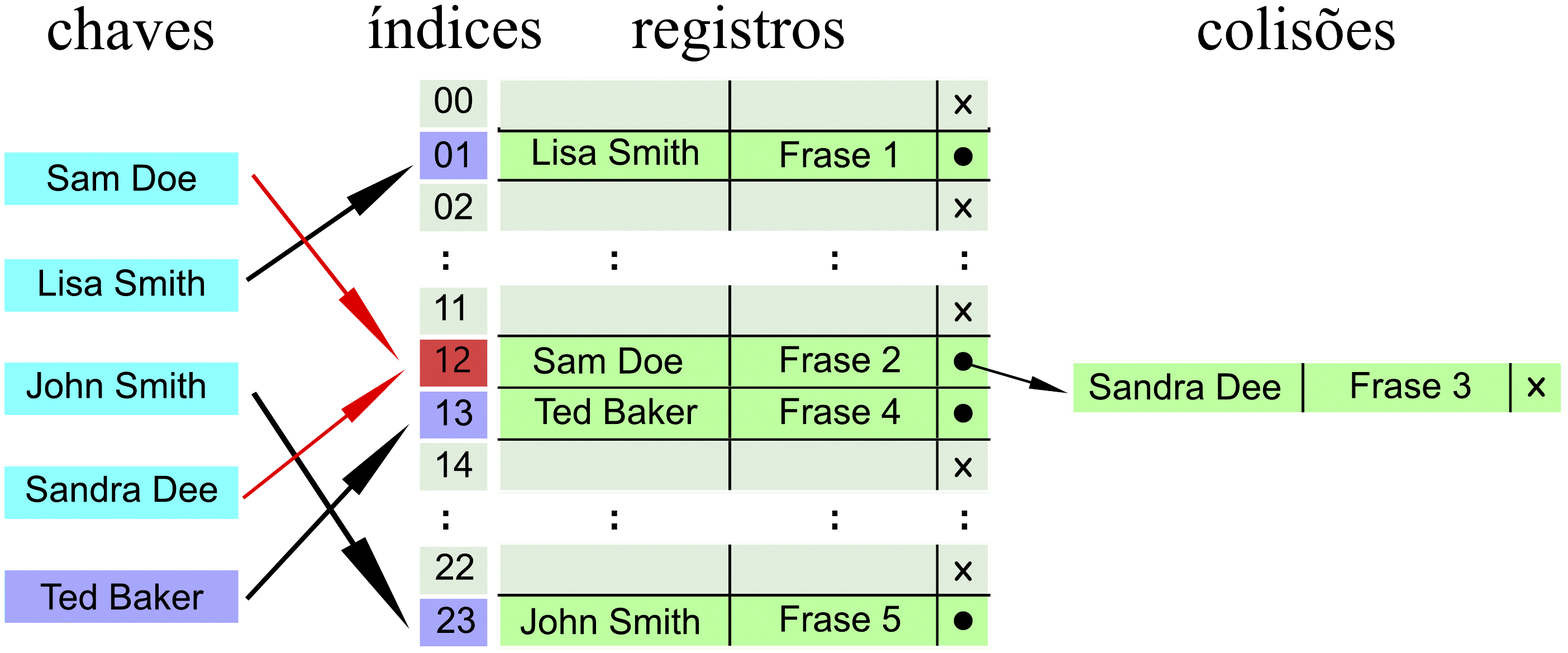}
\caption{Exemplo de uma $Hash Table$ com tratamento de colisão usando listas ligadas.} \label{hash}
\end{figure}

Nosso algortimo para retorno de frases mediante chave única executa os seguintes passos:

\begin{enumerate} 
	\item Geração do índice a partir da chave única fornecida.
	\item  Comparação da chave única informada com a chave única no primeiro elemento no índice obtido para verificar igualdade.
\item Repetição da comparação enquanto não for encontrada uma combinação e ainda houver elementos na lista ligada do índice obtido.  
\item Retorno do ponteiro de caracteres da frase associada a chave única armazenada na $Hash Table$, se não houver uma combinação exata da chave única é retornado o ponteiro para a chave única para facilitar $debug$. 
\end{enumerate}

A inserção de conteúdo na $Hash\ Table$ é constante e o retorno no melhor caso também, no pior caso o retorno será O (n) sendo n a quantidade de colisões no índice onde a chave única está sendo procurada.   
 
O uso de Tabela Hash possui a vantagem de ter uma busca mais rápida para retorna de frases que o uso apenas de uma única estrutura de lista ligada para armazenamento de todas as chaves únicas e frases.

\subsubsection{Pontuação}

Para finalizar o jogo, duelando o "chefe", é necessário obter 10 pontos chamados de pontos de respeito. Inicialmente o jogador possui 1 ponto de respeito e duela contra "servos" do "chefe" a fim de obter mais pontos. Os duelos são partidas que possuem uma determinada quantidade de jogadas, quando o jogador ganha mais da metade das jogadas ele vence a partida obtendo pontos de respeito. Dependendo das características do “servo” o jogador pode ganhar 1, 2 ou 3 pontos numa única partida, ou perder todos os pontos quando perde a partida.

Perdendo todos os pontos de respeito o jogo é finalizado com a derrota do jogador. Obtendo o número necessário de pontos (10 pontos), para duelar o "chefe", o jogador é introduzido a última partida que finalizará o jogo.

A quantidade de pontos de respeito a serem obtidos ou perdidos por "servo", o número máximo de pontos necessários para duelar o "chefe" e a quantidade de jogadas por partida são configuráveis e essas informações podem ser alteradas no arquivo de configuração geral "configuracao.conf" e nos arquivos de configuração dos servos.

\subsubsection{Probabilidade Inimigo}

Enquanto o jogador não obteve os pontos necessários para enfrentar o "chefe" e ainda não perdeu todos os pontos de respeito, ele deve enfrentar "servos". Os "servos" são escolhidos pelo jogo usando pseudo-aleatoriedade e probabilidade.\\

\begin{tabular}{c|c|c|c}
\hline
Servo & Pontos Fornecidos & Pontos Removidos & Probabilidade de ser escolhido \\
\hline
1 & 1 & 1 & 40\%\\
2 & 3 & 2 & 35\%\\
3 & 2 & 3 & 15\%\\
4 & 0 & 10 & 5\%\\
5 & 2 & 3 & 5\%\\
\hline
\end{tabular}

\section{Resultados}

Os resultados obtidos possibilitaram o desenvolvimento de um algorítmo para detecção de padrões pré-estabelecidos de gestos, através da simples análise do comportamento dos pontos da imagem segmentada. Entretando, este tipo de análise, por mais rudimentar que seja apresentou-se eficaz para este problema. 

Ao longo desse trabalho, foi possível obtermos uma visão de alguns dos problemas no campo da visão computacional e da geometria computacional, casos em que demandavam uma maior atenção no desenvolvimento de seus algorítmos, além de cuidado na performance em que algoritmos destes ramos demandam.

\begin{figure}[H]
\centering
\includegraphics[width=.60\textwidth]{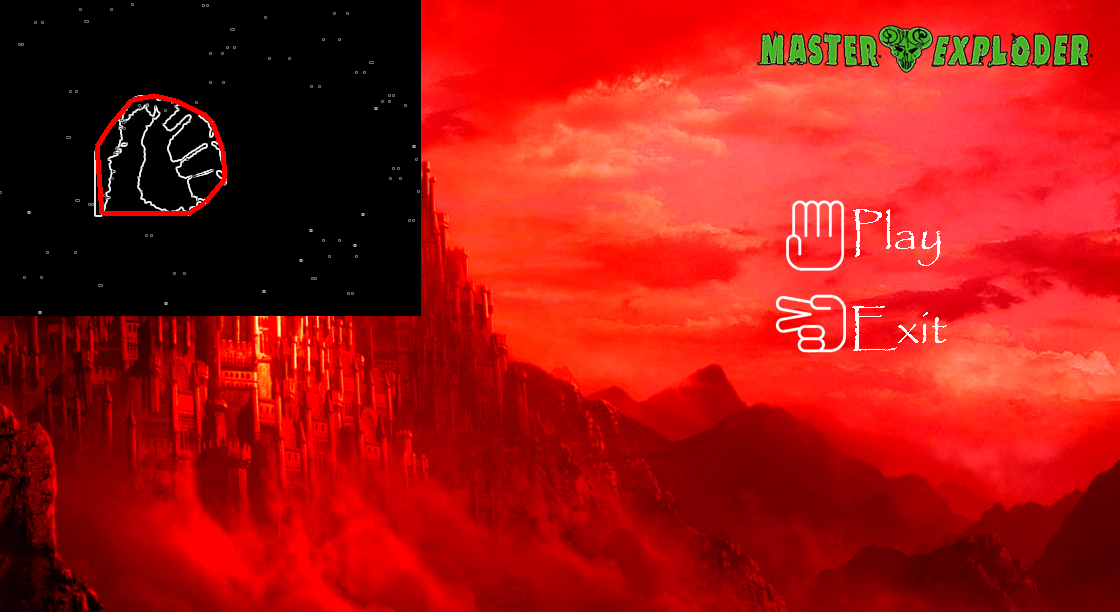}
\caption{Tela de opções com opção de $debug$ ativada nas configurações mostrando no canto superior esquerdo o fecho convexo.} \label{fig1}
\end{figure}

\begin{figure}[H]
\centering
\includegraphics[width=.60\textwidth]{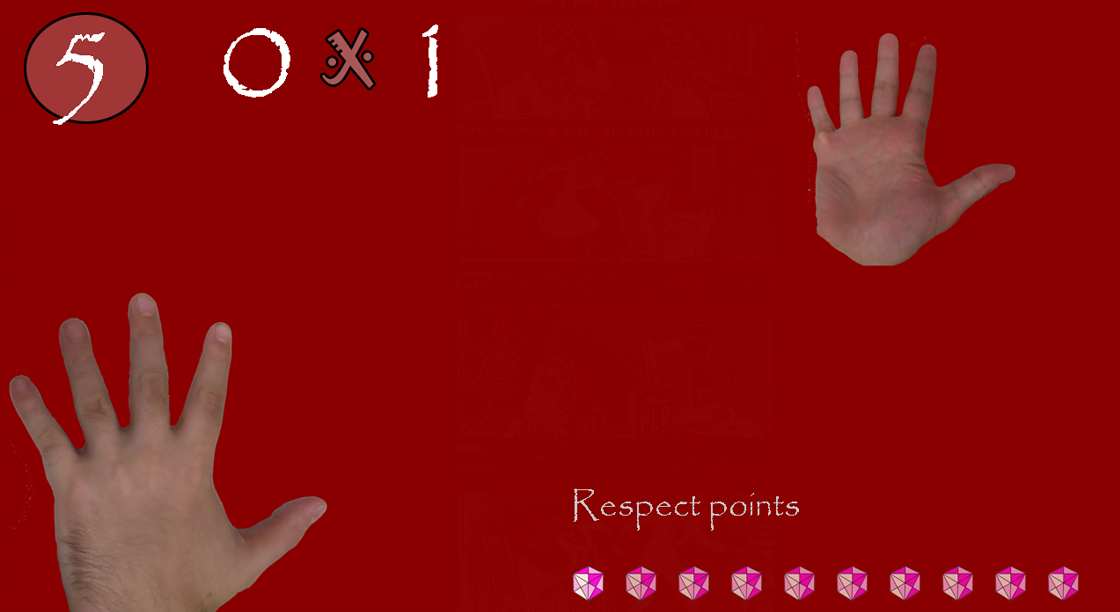}
\caption{Tela da partida. As mãos escolhidas pelo jogador e adversário são reveladas quando o contador no canto superior esquerdo zera.} \label{fig1}
\end{figure}

\section{Conclusão}

O trabalho apresenta como principal contribuição, uma possibilidade de reconhecimento de padrões pré-determinados  para controle de jogos utilizando métodos simples, porém com resultados, dentro de seus limites, precisos.

Os algoritmos propostos são de fácil implementação e não requerem uma abordagem matemática profunda para sua compreensão e aplicação. O trabalho mostra ainda, que estes métodos, com pouca modificação poderiam ser utilizados em qualquer outro tipo de interface por visão computacional, uma vez que seus algoritmos possuem complexidades relativamente médias.

\bibliographystyle{plain}

\end{document}